\documentclass[conference]{IEEEtran}
\IEEEoverridecommandlockouts
\usepackage{cite}
\usepackage{amsmath,amssymb,amsfonts}
\usepackage{algorithmic}
\usepackage{graphicx}
\usepackage{textcomp}
\usepackage{xcolor}
\usepackage{subcaption}
\def\BibTeX{{\rm B\kern-.05em{\sc i\kern-.025em b}\kern-.08em
    T\kern-.1667em\lower.7ex\hbox{E}\kern-.125emX}}
\usepackage[
top = 2.0cm,
bottom  = 2.75cm,
left  = 2.0cm,
right  = 2.0cm]{geometry}

\def\ps@IEEEtitlepagestyle{%
  \def\@oddfoot{\mycopyrightnotice}%
  \def\@evenfoot{}%
}
\def\mycopyrightnotice{%
  {\footnotesize 978-1-7281-6916-3/20/\$31.00~\copyright~2020 IEEE\hfill}% <--- Change here
  \gdef\mycopyrightnotice{}
}

\begin{document}

\title{Low-Resource End-to-end Sanskrit TTS using Tacotron2, WaveGlow and Transfer Learning}
\author{\IEEEauthorblockN{Ankur Debnath}
\IEEEauthorblockA{Department of Electrical Engineering\\
Indian Institute of Science\\
Bengaluru, 560012\\
Email: ankurdebnath@iisc.ac.in}\\
\IEEEauthorblockN{Gangotri Nadiger}
\IEEEauthorblockA{Dept of Information Science
and Engineering\\
SDM College of Engineering and Technology\\
Dharwad, 580002\\
Email: gangotrirg3@gmail.com}\\
\and
\IEEEauthorblockN{Shridevi S Patil}
\IEEEauthorblockA{Dept of Information Science and Engineering\\
SDM College of Engineering and Technology\\
Dharwad, 580002\\
Email: shridevispatil134@gmail.com}\\
\IEEEauthorblockN{Ramakrishnan Angarai Ganesan}
\IEEEauthorblockA{Department of Electrical Engineering\\
Indian Institute of Science\\
Bengaluru, 560012\\
Email: agr@iisc.ac.in}
}

\maketitle

\begin{abstract}
End-to-end text-to-speech (TTS) systems have been developed for European languages like English and Spanish with state-of-the-art speech quality, prosody, and naturalness. However, development of end-to-end TTS for Indian languages is lagging behind in terms of quality. The challenges involved in such a task are: 1) scarcity of quality training data; 2) low efficiency during training and inference; 3) slow convergence in the case of large vocabulary size. In our work reported in this paper, we have investigated the use of fine-tuning the English-pretrained Tacotron2 model with limited Sanskrit data to synthesize natural sounding speech in Sanskrit in low resource settings. Our experiments show encouraging results, achieving an overall MOS of 3.38 from 37 evaluators with good Sanskrit spoken knowledge. This is really a very good result, considering the fact that the speech data we have used is of duration 2.5 hours only.
 
\end{abstract}


\IEEEoverridecommandlockouts
\IEEEpubid{\begin{minipage}[t]{\textwidth}\ \\[10pt]
        \footnotesize{978-1-7281-6916-3/20/\$31.00 \copyright 2020 IEEE}
\end{minipage}} 
%\foot{978-1-7281-6916-3/20/\$31.00 \textcopyright2020 IEEE}

%\cfoot{\tiny{\text 978-1-7281-6916-3/20/\$31.00 \textcopyright2020 IEEE}}
\section{Introduction}
The task of generating natural sounding speech from text remains a challenging problem to be solved. Deep learning based TTS systems are the current state-of-the-art in terms of producing natural sounding speech. Traditionally, concatenative \cite{mileTTS, mileBlizzard2013, mileBlizzard2014} and parametric synthesis techniques were prevalent and had much more complex pipelines and resulted in muffled speech. Besides, the speech output may have glitches and instabilities.

Rapid development in deep learning based methods has shown immense success in this field. End-to-end generative models such as Tacotron2 \cite{Tacotron, Tacotron2} and Deep Voice \cite{Dvoice} have been proposed, which have replaced traditional pipelines. These models have demonstrated state-of-the-art performance by confining the entire pipeline involving spectrogram prediction and speech synthesis into a single pipeline. However, these end-to-end models require tens of hours of speech data and a lot of computational power.

Sanskrit is a heritage language of India and one of the classical languages. Sanskrit has a huge literature and there is a need for producing tools to enable people to listen to ancient Sanskrit works as audios. However, even though there has been a lot of development work on TTS for Indian languages in the past two decades \cite{Thirukkural2001, tts-consortium, FestivalTamil}, there has been hardly any attempt in developing Sanskrit TTS and the only existing work in this regard has been reported by Mishra et al. \cite {mishra}.

However, the main challenge in developing Sanskrit TTS is the lack of sufficient quality data. In this work, we attempt to solve this data-scarcity problem by incorporating fine-tuning or transfer learning \cite{TL-survey, TL_Jia, TL_Fang, TL_Dutoit} from a large pre-trained model available in another language (mainly English). These techniques have proved useful in various applications of deep learning. Also, we perform appropriate data pre-processing that is essential in such low resource settings, in order to achieve reasonably natural sounding speech.

In this paper, we also address the issue of slow convergence and poor training and inference performance arising again due to less data. This is accomplished by appropriately fine tuning the existing English pre-trained Tacotron2 model with our own Sanskrit data, which ensures faster convergence as well as better performance during training and inference.

\begin{figure*}[h]
\begin{center}
     \centerline{\includegraphics[width=0.99\linewidth]{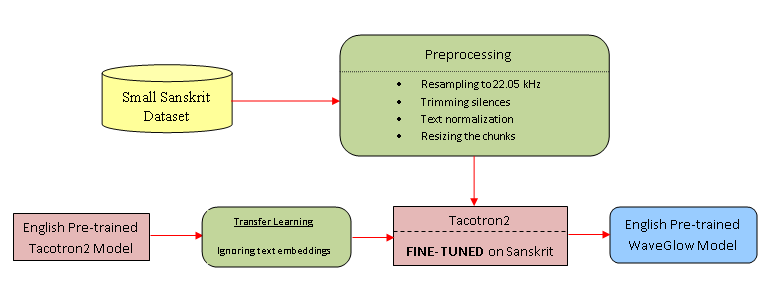}}
    \caption{Block diagram of the designed and developed end-to-end Sanskrit TTS.}
\label{fig:Block}
\end{center}
   
\end{figure*}

\section{Proposed Sanskrit TTS}
A typical neural TTS system is built using multiple stages, but an end-to-end TTS system like Tacotron2 consists of only two major components: (1) A recurrent sequence-to-sequence feature prediction network, which predicts a sequence of mel spectograms from an input character sequence; (2) a vocoder, which generates time-domain waveforms from the sequence of mel spectograms fed to it by the network.

In this paper, the developed Sanskrit TTS uses the Tacotron2\cite{Tacotron2} architecture along with the WaveGlow vocoder\cite{b2}. Figure \ref{fig:Block} gives the block diagram of the entire TTS system. Section \ref{TTS} describes the architecture of the TTS. The details of the dataset collected for the work is described in Sec. \ref{dataset}. Section \ref{pre} discusses the different preprocessing tasks performed. Finally, section \ref{TL} explains the process of transfer learning approach used.

\subsection{End-to-end Neural TTS}\label{TTS}
Among the various deep learning based end-to-end models, Tacotron2\cite{Tacotron2} and WaveGlow\cite{b2} seemed to be appropriate for generating natural sounding speech. The Tacotron2 model is an encoder-attention-decoder setup
where ‘location sensitive attention’ is used. The first part is an encoder which converts the character sequence into word embedding vector. This
representation is later consumed by the decoder to predict spectrograms. To generate time-domain waveforms from the spectograms predicted by the Tacotron2 model, the WaveGlow\cite{b2} vocoder is used.

In this work, we use the Pytorch implementations provided in \cite{taco, wglow}. These two modules need to be trained seperately. The training of the Tacotron2 model and the transfer learning method are discussed in Sec. \ref{TL}. In this work, no explicit training of the vocoder is performed. A pretrained WaveGlow model, trained on the same large English dataset as the Tacotron2, is used. This choice of training procedure can be explained by the fact that the WaveGlow vocoder is fairly consistent over unseen languages and speakers as discussed in \cite{voco}. Also, using a pretrained model reduces computational burden.

\subsection{Sanskrit Speech Data Collection for Training}\label{dataset}
The dataset used is the holy \emph{Bhagavad Gita} and its transcript available online \cite{audio}. It consists of 18 chapters of the holy book in the voice of a male speaker in Sanskrit. This dataset consists of 670 audio clips which are in mp3 format. Transcription is provided for each clip. These audio files and their transcripts have been segmented into chunks of maximum duration 10 sec. The relevant statistics related to the dataset are given in Table \ref{Table3}.

\begin{table}[!htbp]
\centering
\caption{Details of the Sanskrit speech dataset collected}
\resizebox{0.45\textwidth}{0.063\textheight}
{\begin{tabular}{|c|c|}
\hline \textbf{Statistics} & \textbf{Value} \\ \hline
Number of utterances & 1500\\
Total duration & 2h 35min 17sec\\
Vocabulary size & 4888\\
Minimum length of utterance & 2.26 sec\\
Maximum length of utterance & 10.03 sec\\
Average length of utterance & 6.21 sec\\
\hline
\end{tabular}
}
\label{Table3}
\end{table}

\subsection{Data Preprocessing}\label{pre}
An appropriate preprocessing of data prior to training can result in faster convergence and hence better results. In Sec. \ref{res}, we discuss the impact of the preprocessing techniques used, which further validates the facts discussed in \cite{TL_Dutoit}. The main preprocessing strategies adopted in our work are:
\begin{itemize}
    \item Resampling the audio
    \item Trimming the silences
    \item Text normalization and formatting
    \item Resizing the chunks
\end{itemize}

The original audio clips had a sampling rate of 44.1 kHz. However, since both Tacotron2 and WaveGlow were pretrained on LJ speech dataset with a sampling rate of 22.05 kHz, the data needs to be resampled at the same sampling rate to avoid any unwanted glitches.

Silences play a crucial role in a neural TTS system. Especially since Tacotron2 uses a location-sensitive attention, large pauses and silences makes the attention learning slow. Also, a lot of data is required in the presence of large durations of silences. Therefore, trimming of silences in the beginning and end as well as inside silences helps the model converge faster. In this work, any silences beyond 0.5 sec have been removed. This is helpful only if the transcripts do not have any punctuations, because punctuations and silences are learnt simultaneously and both require high amount of data as well as training.

Since the transcripts did not have any abbreviations, numerals etc., no normalization was required. However, to make the system compatible with Devanagari script, the transcripts were converted to unicodes in UTF-8 format, to enable the model train without any issues. Also, it is found that adding end of sentence (EOS) tokens at the end of each utterance further helps the model to converge faster. Hence proper EOS tokens have been added to the transcripts.

The original clips were large recordings and since the pre-trained model was trained on LJ Speech, which has samples of average length 10 secs, we also need to segment the audio and align the corresponding transcripts accordingly, since convergence is harder with samples of larger lengths.

\begin{figure*}[h]
\begin{center}
    \begin{subfigure}{0.45\linewidth}
    \begin{center}
  \includegraphics[width=8.1cm]{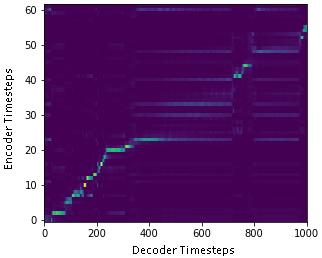}
  
 \caption{Without Preprocessing}   

 \label{fig:subim1}      
    \end{center}
    
\end{subfigure}
\begin{subfigure}{0.45\linewidth}
\begin{center}
    \includegraphics[width=8.1cm]{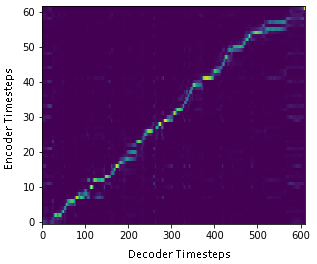}

\caption{With Preprocessing}    

\label{fig:subim2}
\end{center}

\end{subfigure}
 
\caption{Comparison of alignments with and without preprocessing of training data.}
\label{fig:image2}
\end{center}

\end{figure*}

\begin{figure*}[h]
\begin{center}
    \begin{subfigure}{0.45\linewidth}
    \begin{center}
  \includegraphics[width=8cm]{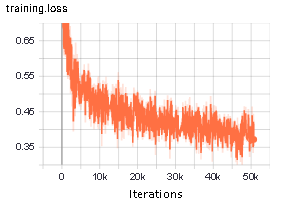} 

 \caption{Training Loss Curve}

\label{fig:subim3}      
    \end{center}

\end{subfigure}
\begin{subfigure}{0.45\linewidth}
\begin{center}
    \includegraphics[width=8cm]{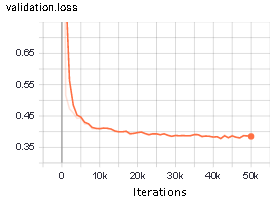}

\caption{Validation Loss Curve}

\label{fig:subim4}
\end{center}

\end{subfigure}
 
\caption{Convergence of the loss as a function of the number of iterations when preprocessing is applied.}
\label{fig:image3}
\end{center}

\end{figure*}

\subsection{Transfer learning from pretrained English model}\label{TL}

In this section, we describe the application of transfer learning and how it is leveraged to fine-tune the pretrained model on the small dataset mentioned in Sec. \ref{dataset}. 

The pre-trained model has been trained on the publicly available LJ Speech dataset, which contains 23.9 hours of single-speaker female speech along with the corresponding transcripts.

Transfer learning from a pretrained model involves selection of the parts of the pretrained model to fine-tune with. This choice depends on the task at hand. In this work, the optimiser details and the text embedding weights are left out and the rest of the pretrained model is used to fine-tune the dataset. Textual information is captured by these embedding weights and is dataset dependent, and is independent of speaker or speaking style. Since the pretrained model was trained on English, it is obvious to ignore them and fine-tune only the speech properties of the pretrained model. Transfer learning in low resource settings has significant impact on convergence of the model when compared with random initialization of the model parameters.

In the case of the vocoder, as explained in Sec. \ref{TTS}, the pretrained WaveGlow model is used since it is used only during inference and being robust to gender and language modifications, using it directly seems to be a justified choice. This reduces computational burden as well as the total training time.

\section{Experiments and Results}\label{res}

Evaluation experiments with the proposed system are conducted in two steps. First, transfer learning with and without preprocessing are compared against each other in terms of training losses and alignment. Also, experiments are carried out by varying different network hyperparameters such as gate threshold, decoder and attention layer dropout probabilities and optimization hyperparameters such as batch size, learning rate and weight decay of the Adam optimizer used in the model. These experiments were carried out on a GPU server consisting of 4 NVIDIA RTX 2080Ti of 12 GB capacity each. It is to be noted that training of Tacotron2 with random initialization was not attempted taking into account the fact that it will not learn anything at all given the size of the data used. The reader is suggested to read \cite{TL_Dutoit} for further details.

The second part of the experiment involves evaluating the naturalness and accuracy of the synthesized speech output in terms of mean opinion score (MOS).

\subsection{Model Evaluation}

Initially, the Tacotron2 model is trained with the dataset mentioned in Sec. \ref{dataset} without the preprocessing mentioned in Sec. \ref{pre}. It is found that the model is unable to converge and learn proper alignment even after training the model up to 30,000 iterations. Even after a lot of hyperparameter tuning, the results are not acceptable.

A second set of experiments are conducted, now with the proper preprocessing mentioned in Sec. \ref{pre}. Finally after proper tuning of the hyperparameters and training the model up to 50,000 iterations, the synthesized outputs are encouraging even for unseen texts. The final set of hyperparameters are given in Tables \ref{Table1} and \ref{Table2}.

\begin{table}[!htbp]
\centering
\caption{Final values of the model hyperparameters}
\resizebox{0.3\textwidth}{0.06\textheight}
{
\begin{tabular}{|c|c|}
\hline \textbf{Hyperparameter} & \textbf{Value} \\ \hline
Epochs & 430\\
Batch size & 8\\
Gate threshold & 0.4\\
Decoder dropout & 0.4\\
Attention dropout & 0.4\\
\hline
\end{tabular}
}
\label{Table1}
\end{table}

\begin{table}[!htbp]
\centering
\caption{Adam Optimizer hyperparameter values used}
\resizebox{0.33\textwidth}{0.056\textheight}
{\begin{tabular}{|c|c|}
\hline \textbf{Optimizer parameters} & \textbf{Value} \\ \hline
Learning rate, $\eta$ & $4*10^{-5}$ \\
$\beta_{1}$ & 0.9\\
$\beta_{2}$ & 0.999\\
Weight decay, $\epsilon$ & $10^{-5}$\\
\hline
\end{tabular}
}
\label{Table2}

\end{table}

From these experiments, it is found that annealing the learning rate helps the model converge faster. Also, increasing dropout probabilities in low resource environments helps in learning attention properly.

Figure \ref{fig:image2} compares the learned alignments with and without preprocessing, with all the other hyperparameters same as Tables \ref{Table1} and \ref{Table2}. In Fig. \ref{fig:image2}, the alignment plots are shown for both cases. An alignment plot shows how well the model has been able to learn the attention. A diagonal or a near-perfect diagonal indicates perfect alignment and generates natural sounding speech\cite{Tacotron2}. It can be seen in Fig. \ref{fig:subim1}, that without preprocessing, the model fails to learn attention properly. Whereas, in Fig. \ref{fig:subim2}, the model learns the attention properly. The possible explanation for this difference is that in scenarios where data is scarce, removing all the silences (in the beginning, inside, and end of sentences) helps the model learn attention properly and converge faster resulting in clear and natural sounding speech output. Figure \ref{fig:image3} shows the loss curves of the final training.

\subsection{Output Evaluation}

The quality of the synthesized speech was evaluated by 37 people, who are well-versed in Sanskrit and can read and write in Sanskrit. Each of them evaluated the output utterance separately on naturalness and accuracy of pronunciation. They were made familiar with the MOS scale before being asked to evaluate the synthesized sentence \cite{tts-eval}.
The results of this experiment are given in Table \ref{Table4}.We have obtained a MOS of $3.41 \pm 0.33$ for naturalness and $3.35 \pm 0.29$ for accuracy of pronunciation considering a confidence interval of $95\%$, which we consider are very good, given the very limited training data of only 2.5 hours duration. This has been made possible primarily due to transfer learning.

\begin{table}[!htbp]
\centering
\caption{Evaluation of the quality of the synthesized output by 37 volunteers with good knowledge of Sanskrit, with 95\% confidence intervals computed from the t-distribution for various systems.}
\resizebox{0.45\textwidth}{0.048\textheight}
{
\begin{tabular}{|c|c|}
\hline \textbf{Characteristic evaluated}  & \textbf{Mean opinion score} \\ \hline
Naturalness & $3.41 \pm 0.33$\\
Accuracy of pronunciation & $3.35 \pm 0.29$\\ 

\hline
Overall & $3.38 \pm 0.31$\\
\hline
\end{tabular}
}
\label{Table4}
\end{table}

\section{Conclusion and Future work}

In this paper, we design, develop and test an end-to-end TTS for Sanskrit in a low resource environment. The problem of data-scarcity is dealt with using transfer learning i.e. fine-tuning a large pretrained Tacotron2 model using the available small Sanskrit speech dataset. We also address the issue of slow convergence and poor training and inference performance by applying appropriate data preprocessing techniques as well as suggestive hyperparameter tuning for achieving faster learning of attention.

In future, the proposed system can be further improved by
exploring transfer learning among similar languages, which have comparable number of phonemes. We also think that in order to synthesize code-mixed speech, it may be preferable to use grapheme-to-phoneme converters \cite{TamilG2P} for both the input languages and use phoneme sequence as the input to the Tacotron2 architecture, rather than the character sequence. Also, there is scope for significant research in terms of incorporating pretrained language models\cite{TL_Fang} to further improve the training performance and achieve better alignment in similar resource settings.

\section*{Acknowledgment}
The authors thank Dr. Ramanujachar for pointing to sources for downloading recorded Sanskrit chants.

\end{document}